# Inferring Latent Structure From Mixed Real and Categorical Relational Data

**Esther Salazar**[1]                                   ESTHER.SALAZAR@DUKE.EDU
**Matthew S. Cain**[2]                              MATTHEW.S.CAIN@DUKE.EDU
**Elise F. Darling**[2]                                           ED75@DUKE.EDU
**Stephen R. Mitroff**[2]                                  MITROFF@DUKE.EDU
**Lawrence Carin**[1]                                        LCARIN@DUKE.EDU

[1]Department of Electrical & Computer Engineering, Duke University, Durham, NC, USA
[2]Department of Psychology and Neuroscience, Duke University, Durham, NC, USA

## Abstract

We consider analysis of relational data (a matrix), in which the rows correspond to subjects (e.g., people) and the columns correspond to attributes. The elements of the matrix may be a mix of real and categorical. Each subject and attribute is characterized by a latent binary feature vector, and an inferred matrix maps each row-column pair of binary feature vectors to an observed matrix element. The latent binary features of the rows are modeled via a multivariate Gaussian distribution with low-rank covariance matrix, and the Gaussian random variables are mapped to latent binary features via a probit link. The same type construction is applied *jointly* to the columns. The model infers latent, low-dimensional binary features associated with each row and each column, as well correlation structure between all rows and between all columns.

## 1. Introduction

The inference of low-dimensional latent structure in matrix and tensor data constitutes a problem of increasing interest. For example, there has been a significant focus on exploiting low-rank and related structure in many types of matrices, primarily for matrix completion (Lawrence & Urtasun, 2009; Yu et al., 2009; Salakhutdinov & Mnih, 2008). In that problem one is typically given a very small fraction of the total matrix, and the goal is to infer the missing entries. In other problems, all or most of the matrix is given, and the goal is to infer relationships between the rows, and between the columns. For that problem, co-clustering has received significant attention (Dhillon et al., 2003; Wang et al., 2011). In co-clustering the rows/columns are typically mapped to hierarchical clusters, which may be overly restrictive in some cases. Specifically, there are situations for which two or more rows/columns may have a subset of (latent) characteristics in common, but differ in other respects, and therefore explicit assignment to clusters is inappropriate. This motivates so-called mixed-membership models. For instance, in (Meeds et al., 2007) the authors develop a model in which each row and column has an associated binary feature vector, representing each in terms of the presence/absence of particular latent features. Rather than explicitly assigning cluster membership, the binary features assign "mixed memberships," because rows/columns may partially share particular latent features. In (Meeds et al., 2007) the latent binary features are mapped to observed matrix elements via an intervening regression matrix, which is also inferred. Rather than using binary features to represent the rows and columns, one may also use a sparse *real* feature vector for each row and column (Salakhutdinov & Mnih, 2008; Wang et al., 2010), as is effectively done in factor analysis (Carvalho et al., 2008). As noted in (Meeds et al., 2007), and discussed further below, the use of binary feature vectors aids model interpretation, and may also enhance data sharing.

The Indian buffet process (IBP) (Griffiths & Ghahramani, 2005) is a natural tool for modeling latent binary feature vectors, and that approach was taken in (Meeds et al., 2007). The IBP is closely related to the beta-Bernoulli process (Thibaux & Jordan, 2007), which implies that each row (subject) effectively selects a given feature i.i.d. from an underlying Bernoulli distribution, with feature-dependent

---

(but subject-independent) Bernoulli probability. It is expected that many subjects may be closely related, and therefore these are likely to have similar latent binary features, with the same expected of the columns. This statistical correlation between subsets of rows/columns motivates the co-clustering approaches, but for reasons stated above clustering is often too restrictive.

To address these limitations of existing approaches, we propose a new model, in which the rows/columns are each characterized by latent binary feature vectors, as in (Meeds et al., 2007). However, a new construction is used to model the binary row/column features, moving beyond the i.i.d. assumption that underlies the IBP. Assume the matrix of interest is characterized by $N$ rows. For each of the latent features of the rows, an $N$-dimensional real vector is drawn from a zero-mean multivariate Gaussian distribution, with an unknown covariance structure. This $N$-dimensional real vector is employed with a probit link function to constitute an $N$-dimensional binary vector, manifesting the row-dependent binary value for one of the latent features. This is done for all latent binary features. By inferring the underlying covariance matrix, we uncover the latent correlation between the rows, without explicit clustering. Multivariate probit models are utilized jointly for the simultaneous analysis of rows and columns; this embodies the joint analysis of rows and columns manifested by co-clustering, while performing such in a mixed-membership setting.

For large $N$ (*i.e.*, many rows or columns), one must impose structure on the row/column covariance matrices, to achieve computational efficiency, and to enhance the uncovering of structure. This is implemented by imposing a low-rank covariance matrix model, via factor analysis with sparse factor loadings (Carvalho et al., 2008). The rows (columns) that share non-zero values in the factor loadings are inferred to be statistically correlated, without the necessity of imposing explicit clustering.

Bayesian model inference is performed, via efficient MCMC. The model is demonstrated on three real datasets.

## 2. Basic Model Setup

### 2.1. Problem statement

We consider data from $N$ subjects, with the data in general a mix of categorical and real. There are $M_1$ categorical entries and $M_2$ real entries. The categorical data are represented as an $N \times M_1$ matrix $\boldsymbol{X}$, and the real entries are represented by the $N \times M_2$ matrix $\boldsymbol{Y}$;

we wish to analyze $\boldsymbol{X}$ and $\boldsymbol{Y}$ jointly.

The transpose of the column vector $\boldsymbol{x}_i$ represents the $i$th row of $\boldsymbol{X}$, and the transpose of the column vector $\boldsymbol{y}_i$ represents the $i$th row of $\boldsymbol{Y}$. Vector $\boldsymbol{x}_i = (x_{i1}, \ldots, x_{iM_1})^T$ contains categorical observations where $x_{ij} \in \{0, \ldots, q_j - 1\}$ and $q_j$ corresponds to the number of categories associated with the $j$th component.

### 2.2. Factor analysis

A $q_j$-dimensional probit-regression model is employed for $x_{ij}$. Specifically, assume that there is a feature vector $\boldsymbol{v}_i \in \mathbb{R}^{K_x}$ associated with subject $i$ (as discussed below, the need to set $K_x$ disappears in the final form of the model). The observed multinomial variable $x_{ij}$ is modeled in terms of a latent variable $\boldsymbol{\beta}_{ij} \in \mathbb{R}^{q_j - 1}$ such that

$$\boldsymbol{\beta}_{ij} = \boldsymbol{S}_j^T \boldsymbol{v}_i + \boldsymbol{\epsilon}_{ij}, \qquad \boldsymbol{\epsilon}_{ij} \sim \mathcal{N}(0, \boldsymbol{\Sigma}_j)$$

$$x_{ij} = \begin{cases} 0 & \text{if} \quad \max_{1 \le l \le q_j - 1} \beta_{ij}^{(l)} < 0 \\ p & \text{if} \quad \max_{1 \le l \le q_j - 1} \beta_{ij}^{(l)} = \beta_{ij}^{(p)} > 0 \end{cases}$$

where $p = 1, \ldots, q_j - 1$, $\boldsymbol{S}_j = (\boldsymbol{s}_j^{(1)}, \ldots, \boldsymbol{s}_j^{(q_j-1)}) \in \mathbb{R}^{K_x \times q_j - 1}$, $\boldsymbol{s}_j^{(p)} \in \mathbb{R}^{K_x}$, $\{\boldsymbol{\Sigma}_j\}_{11} = 1$ and $x_{ij} = 0$ corresponds to the base category. Note that $\boldsymbol{\Sigma}_j$ is a $(q_j - 1) \times (q_j - 1)$ covariance matrix, and the first element of $\boldsymbol{\Sigma}_j$ is fixed to 1 in order to avoid identifiability problems (Chib & Greenberg, 1998; Zhang et al., 2008). The covariance matrix $\boldsymbol{\Sigma}_j$ infers statistical correlation between the $q_j$ possible categories associated with attribute $j$, and an inverse-Wishart prior is employed for $\boldsymbol{\Sigma}_j$.

The $p$th component of $\boldsymbol{\beta}_{ij}$ is given by

$$\beta_{ij}^{(p)} = \boldsymbol{v}_i^T \boldsymbol{s}_j^{(p)} + \epsilon_{ij}^{(p)}, \quad p = 1, \ldots, q_j - 1, \quad (1)$$

where $\epsilon_{ij}^{(p)} \sim \mathcal{N}(0, \{\boldsymbol{\Sigma}_j\}_{pp})$ and $\boldsymbol{s}_j^{(p)} \in \mathbb{R}^{K_x}$ represents a feature vector associated with the choice $p \in \{1, \ldots, q_j - 1\}$ for the component $j$ of $\boldsymbol{x}_i$.

A similar construction is employed for the components of the real matrix $\boldsymbol{Y}$, without the need for the probit link. Specifically, for row $i$ and column $j$ we respectively define real feature vectors $\boldsymbol{a}_i \in \mathbb{R}^{K_y}$ and $\boldsymbol{b}_j \in \mathbb{R}^{K_y}$, and the $(i, j)$th matrix entry is modeled as

$$y_{ij} = \boldsymbol{a}_i^T \boldsymbol{b}_j + \varepsilon_{ij}, \quad \varepsilon_{ij} \sim N(0, \sigma_y^2). \quad (2)$$

with $K_y$ again disappearing in the final form of the model.

### 2.3. Binary row and column feature vectors

Let $\boldsymbol{r}_i \in \{0, 1\}^K$ represent a latent *binary* feature vector characteristic of row $i$ in both $\boldsymbol{X}$ and $\boldsymbol{Y}$. We assu-

me that

$$\boldsymbol{v}_i = \boldsymbol{R}^{(X)}\boldsymbol{r}_i \quad \text{and} \quad \boldsymbol{a}_i = \boldsymbol{R}^{(Y)}\boldsymbol{r}_i, \quad (3)$$

where $\boldsymbol{R}^{(X)} \in \mathbb{R}^{K_x \times K}$, $\boldsymbol{R}^{(Y)} \in \mathbb{R}^{K_y \times K}$. Similarly, let $\boldsymbol{d}_j^{(p)} \in \{0,1\}^K$ represent the latent *binary* feature vector associated with $\boldsymbol{s}_j^{(p)}$, with $\boldsymbol{c}_j \in \{0,1\}^K$ so defined for $\boldsymbol{b}_j$ (for notational simplicity we write all binary vectors as being of same dimension $K$, but in practice the model infers the number of binary components needed to represent each of these vectors). We assume

$$\boldsymbol{s}_j^{(p)} = \boldsymbol{C}^{(X)}\boldsymbol{d}_j^{(p)} \quad \text{and} \quad \boldsymbol{b}_j = \boldsymbol{C}^{(Y)}\boldsymbol{c}_j, \quad (4)$$

where $\boldsymbol{C}^{(X)} \in \mathbb{R}^{K_x \times K}$ and $\boldsymbol{C}^{(Y)} \in \mathbb{R}^{K_y \times K}$. Using the above constructions, we can rewrite (1) and (2) as

$$\begin{aligned}\beta_{ij}^{(p)} &= \boldsymbol{r}_i^T \boldsymbol{M}^{(X)} \boldsymbol{d}_j^{(p)} + \epsilon_{ij}^{(p)}, & (5)\\ y_{ij} &= \boldsymbol{r}_i^T \boldsymbol{M}^{(Y)} \boldsymbol{c}_j + \varepsilon_{ij}. & (6)\end{aligned}$$

Note that the need to set the aforementioned dimensions $K_x$ and $K_y$ has been removed, and what remains are the matrices $\boldsymbol{M}^{(X)} = \boldsymbol{R}^{(X)^T} \boldsymbol{C}^{(X)} \in \mathbb{R}^{K \times K}$ and $\boldsymbol{M}^{(Y)} = \boldsymbol{R}^{(Y)^T} \boldsymbol{C}^{(Y)} \in \mathbb{R}^{K \times K}$; $K$ is typically set as a large upper bound on the number of binary features needed to represent the rows and columns, with only a subset of these $K$ variables inferred as important when performing computations.

The advantage of this construction is that real feature vectors $\{\boldsymbol{v}_i\}$, $\{\boldsymbol{s}_j^{(p)}\}$, $\{\boldsymbol{a}_i\}$ and $\{\boldsymbol{b}_i\}$ are constituted in terms of *binary* feature vectors, with the regression matrices $\boldsymbol{M}^{(X)}$ and $\boldsymbol{M}^{(Y)}$ between the binary and real vectors *shared* for all rows and columns. This imposes significant structure and sharing on the learning of $\{\boldsymbol{v}_i\}$, $\{\boldsymbol{s}_j^{(p)}\}$, $\{\boldsymbol{a}_i\}$ and $\{\boldsymbol{b}_i\}$, as considered in (Meeds et al., 2007). This paper differs from (Meeds et al., 2007) in three ways: (*i*) we jointly consider real and categorical data jointly, (*ii*) we impose low-rank structure on $\boldsymbol{M}^{(X)}$ and $\boldsymbol{M}^{(Y)}$ (discussed next), and (*ii*) a new framework is developed for modeling the binary vectors $\{\boldsymbol{r}_i\}$, $\{\boldsymbol{d}_j^{(p)}\}$ and $\{\boldsymbol{c}_j\}$ (discussed in Section 3).

### 2.4. Low-rank regression matrices

We model $\boldsymbol{M}^{(X)}$ and $\boldsymbol{M}^{(Y)}$ as low-rank matrices:

$$\boldsymbol{M}^{(X)} = \sum_{l=1}^K \lambda_l^{(X)} b_l^{(X)} \boldsymbol{u}_l^{(X)} (\boldsymbol{v}_l^{(X)})^T, \quad (7)$$

$$\boldsymbol{M}^{(Y)} = \sum_{l=1}^K \lambda_l^{(Y)} b_l^{(Y)} \boldsymbol{u}_l^{(Y)} (\boldsymbol{v}_l^{(Y)})^T, \quad (8)$$

with $\lambda_l^{(X)} \sim \mathcal{N}_{0,\infty}(0, \sigma_\lambda^2)$, corresponding to a truncated normal distribution, over $(0, \infty)$, with an inverse-gamma prior on $\sigma_\lambda^2$; $\lambda_l^{(Y)}$ is defined similarly. The vectors $\boldsymbol{u}_l^{(X)}$, $\boldsymbol{v}_l^{(X)}$, $\boldsymbol{u}_l^{(Y)}$ and $\boldsymbol{v}_l^{(Y)}$ are all defined similarly, and we discuss one in detail, for conciseness. Specifically, we draw $\boldsymbol{u}_l^{(X)} \sim \mathcal{N}(0, \boldsymbol{I}_K)$, where $\boldsymbol{I}_K$ is the $K \times K$ identity matrix. The variables $b_l^{(X)}$ and $b_l^{(Y)}$ are binary, and they allow inference of the associated matrix rank. Again illustrating one of these for conciseness, we employ a sparseness-inducing beta-Bernoulli representation, with $b_l^{(X)} \sim \text{Bernoulli}(\pi^{(X)})$, with $\pi^{(X)} \sim \text{Beta}(1/K, 1)$.

Using, for example, (7) in (5), we observe that the model imposes

$$\beta_{ij}^{(p)} = \sum_{l \in \mathcal{S}} \lambda_l^{(X)} < \boldsymbol{r}_i, \boldsymbol{u}_l^{(X)} > < \boldsymbol{d}_j^{(p)}, \boldsymbol{v}_l^{(X)} > + \epsilon_{ij}^{(p)} \quad (9)$$

where $< \cdot, \cdot >$ corresponds to a vector inner product, and $\mathcal{S}$ defines the set of indices for which $b_l^{(X)} = 1$. Note that for the probit link function this construction implies that we do not require random-effect terms for the subjects and attributes (as is typically done when using real feature vectors (Wang et al., 2010)), since the sum of terms in (9) automatically allow random-effect terms, if needed (such will correspond to one of the terms in the sum).

From (9), and considering (1), note that we may represent $\boldsymbol{v}_i$ as a vector composed of $\sqrt{\lambda_l^{(X)}} < \boldsymbol{r}_i, \boldsymbol{u}_l^{(X)} >$ for $l \in \mathcal{S}$; $\boldsymbol{s}_j^{(p)}$ is similarly defined by $\sqrt{\lambda_l^{(X)}} < \boldsymbol{d}_j^{(p)}, \boldsymbol{v}_l^{(X)} >$ for $l \in \mathcal{S}$. Therefore, via the low-rank construction in (7), we *infer* $K_x$ to be the size of set $\mathcal{S}$ (rank of $\boldsymbol{M}^{(X)}$). As discussed when presenting results, the low-rank construction in (8) allows us to similarly infer $K_y$. This property is a principal reason for imposing low-rank structure on $\boldsymbol{M}^{(X)}$ and $\boldsymbol{M}^{(Y)}$, it aiding interpretation of the model results.

## 3. Correlated Binary Feature Vectors

We describe in detail the proposed modeling of $\boldsymbol{r}_i$, with a similar construction employed for $\{\boldsymbol{d}_j^{(p)}\}$ and $\{\boldsymbol{c}_j\}$. A sparse *multivariate* probit model is imposed: $\boldsymbol{\eta}_k \sim \mathcal{N}(0, \boldsymbol{\Sigma})$, with $r_{ik}|\eta_{ik} = 1$ if $\eta_{ik} > 0$, and $r_{ik}|\eta_{ik} = 0$ otherwise; $k = 1, \ldots, K$, with $\eta_{ik}$ the $i$th component of $\boldsymbol{\eta}_k$, and $r_{ik}$ is the $k$th component of $\boldsymbol{r}_i$. Marginally, $r_{ik} \sim \text{Ber}(\pi_{ik})$ where $\pi_{ik} = \text{Pr}(\eta_{ik} > 0)$. The covariance matrix $\boldsymbol{\Sigma} \in \mathbb{R}^{N \times N}$ imposes an underlying correlation structure between $(r_{1k}, \ldots, r_{Nk})$, for all binary features $k$.

We must now place a prior on the covariance matrix $\boldsymbol{\Sigma}$. A large class of models impose sparsity on the inverse of $\boldsymbol{\Sigma}$ (*i.e.*, the precision matrix), corresponding to a sparse Gaussian graphical models (GGM). The GGM approach for covariance matrix estimation is attrac-tive and many approaches have been proposed (Atay-Kayis & Massam, 2005; Dobra et al., 2011). Alternatively, other approaches have been proposed for directly modeling the covariance matrix, placing shrinkage priors on various parameterizations of $\boldsymbol{\Sigma}$. For instance, (Liechty et al., 2004) considered shrinkage priors in terms of the correlation matrix, and (Yang & Berger, 1994) used reference priors based on the spectral decomposition of $\boldsymbol{\Sigma}$.

We choose to directly model the components of $\boldsymbol{\Sigma}$. In fact, given that $N$ (dimensionality of the covariance matrix) is typically larger than $K$, standard estimators are liable to be unstable (Sun & Berger, 2006; Hahn et al., 2012). Hence, we impose a factor structure on a covariance matrix, in a similar fashion to (Hahn et al., 2012). Such regularization is crucial when the number of variables is large relative to the sample size, and also when the covariance corresponds to an unobservable latent variable (Rajaratman et al., 2008).

We assume that $\mathrm{cov}(\boldsymbol{\eta}_k) = \boldsymbol{B}\boldsymbol{B}^T + \boldsymbol{\Psi}$ where $\boldsymbol{\Psi} \in \mathbb{R}^{N \times N}$ is a diagonal matrix, $\mathrm{rank}(\boldsymbol{B}) = K < N$ and $\boldsymbol{B} \in \mathbb{R}^{N \times K}$. That construction implies that $\boldsymbol{\eta}_k \sim \mathcal{N}(\boldsymbol{B}\boldsymbol{f}_k, \boldsymbol{\Psi})$ where $\boldsymbol{f}_k \sim \mathcal{N}(0, \boldsymbol{I}_K)$. The matrix $\boldsymbol{B}$ must be constrained to be zero for upper-triangular entries and positive along the diagonal, to avoid identifiability problems. Further, $\boldsymbol{\Psi}$ is fixed to be the identity to allow a simple identification strategy. The prior on the loadings $\boldsymbol{B}$ is given by

$$(b_{ik}|v_i, \pi_k) \sim \pi_k \mathcal{N}(0, v_i) + (1 - \pi_k)\delta_0 \quad (10)$$

with $v_i \sim \mathrm{IG}(c/2, cd/2)$ and $\pi_i \sim \mathrm{Beta}(1,1)$, with IG an inverse-gamma distribution and $\delta_0$ a unit point measure concentrated at 0. The sparsity prior permits some of the unconstrained elements in the factor-loadings matrix $\boldsymbol{B}$ to be identically zero.

## 4. Posterior Computation

An approximation to the full posterior of model parameters is performed based on a Gibbs sampler, with Metropolis-Hastings updates for a subset of the parameters. We now briefly describe how to sample some of the most interesting parameters, based on their full conditional posterior distributions.

• Sample $\boldsymbol{u}_l^{(X)}$ as $(\boldsymbol{u}_l^{(X)}|-) \sim \mathcal{N}(\boldsymbol{m}_l^{(X)}, \boldsymbol{V}_l^{(X)})$, where $\boldsymbol{V}_l^{(X)} = \left(\boldsymbol{I}_K + \sum_{i=1}^{N}(\tilde{\boldsymbol{E}}_i^{(l)})^T \tilde{\boldsymbol{\Sigma}}^{-1} \tilde{\boldsymbol{E}}_i^{(l)}\right)^{-1}$, $\boldsymbol{m}_l^{(X)} = \boldsymbol{V}_l^{(X)} \sum_{i=1}^{N} (\tilde{\boldsymbol{E}}_i^{(l)})^T \tilde{\boldsymbol{\Sigma}}^{-1} \boldsymbol{\beta}_i^{(-l)}$, $\tilde{\boldsymbol{E}}_i^{(l)} = \lambda_i^{(X)} \boldsymbol{D}^T \boldsymbol{v}_l^{(X)} \boldsymbol{r}_i^T$, $\boldsymbol{D} = (\boldsymbol{d}_1, \ldots, \boldsymbol{d}_{M_1})$, $\boldsymbol{d}_j = (\boldsymbol{d}_j^{(1)}, \ldots, \boldsymbol{d}_j^{(q_{j-1})})$, $\tilde{\boldsymbol{\Sigma}} = \mathrm{diag}(\boldsymbol{\Sigma}_1, \ldots, \boldsymbol{\Sigma}_{M_1})$, $\boldsymbol{\beta}_i^{(-l)} = \boldsymbol{\beta}_i - \boldsymbol{D}^T (\sum_{k \neq l} \lambda_k^{(X)} b_k^{(X)} \boldsymbol{v}_k^{(X)} (\boldsymbol{u}_k^{(X)})^T) \boldsymbol{r}_i$, and $\boldsymbol{\beta}_i^T = (\boldsymbol{\beta}_{i1}^T, \ldots, \boldsymbol{\beta}_{iM_1}^T) \in \mathbb{R}^{q_1 + \ldots + q_{M_1} - M_1}$.

• In order to sample $r_{ik} \in \{0,1\}$, $i = 1, \ldots, N$, $k = 1, \ldots, K$, let $\boldsymbol{y}_i = \tilde{\boldsymbol{M}}_Y \boldsymbol{r}_i + \boldsymbol{\epsilon}_i$ and $\boldsymbol{\beta}_i = \tilde{\boldsymbol{M}}_X \boldsymbol{r}_i + \boldsymbol{\varepsilon}_i$ where $\tilde{\boldsymbol{M}}_Y = \boldsymbol{C}^T \boldsymbol{M}^{(Y)^T}$ and $\tilde{\boldsymbol{M}}_X = \boldsymbol{D}^T \boldsymbol{M}^{(X)^T}$. Also, let $\boldsymbol{y}_i^{(-k)} = \boldsymbol{y}_i - \tilde{\boldsymbol{M}}_Y^{(-k)} \boldsymbol{r}_i^{(-k)}$ and $\boldsymbol{\beta}_i^{(-k)} = \boldsymbol{\beta}_i - \tilde{\boldsymbol{M}}_X^{(-k)} \boldsymbol{r}_i^{(-k)}$. Then, $(r_{ik}|-) \sim \mathrm{Bernoulli}(p_1/(p_1 + p_2))$, where $p_1 = \pi_{ik} \exp\{-0.5(\sigma_y^{-2} \boldsymbol{e}_i^T \boldsymbol{e}_i + \boldsymbol{f}_i^T \tilde{\boldsymbol{\Sigma}}^{-1} \boldsymbol{f}_i)\}$, $p_2 = (1 - \pi_{ik}) \exp\{-0.5(\sigma_y^{-2} \boldsymbol{y}_i^{(-k)^T} \boldsymbol{y}_i^{(-k)} + \boldsymbol{\beta}_i^{(-k)^T} \tilde{\boldsymbol{\Sigma}}^{-1} \boldsymbol{\beta}_i^{(-k)})\}$, $\boldsymbol{e}_i = \boldsymbol{y}_i^{(-k)} - \boldsymbol{M}_Y^{(-k)}$, $\boldsymbol{f}_i = \boldsymbol{\beta}_i^{(-k)} - \boldsymbol{M}_X^{(-k)}$ and $\pi_{ik} = \Pr(\eta_{ik} > 0)$. Samples for $c_{jk}$, $d_{jk}^{(p)}$, $b_l^{(X)}$ and $b_l^{(Y)}$ are obtained similarly.

• The parameter-extended Metropolis-Hastings algorithm is employed to sample $\boldsymbol{\Sigma}_j$ given the restriction $\{\boldsymbol{\Sigma}_j\}_{11} = 1$ (Zhang et al., 2008) only when $q_j > 2$, otherwise $\boldsymbol{\Sigma}_j$ is fixed to one. Considering a Wishart prior $\boldsymbol{\Sigma}_j \sim \mathrm{W}(m_0, \Omega_j)$, the algorithm is as follows: (1) at iteration $t$, set the values $(R_j^{(t)}, D_j^{(t)})$ by generating $\boldsymbol{\Sigma}_j = D_j^{(t)1/2} R_j^{(t)} D_j^{(t)1/2}$, where $R_j^{(t)}$ is the correlation matrix and $D_j^{(t)}$ the diagonal variance matrix with the first element equal to one. (2) Generate candidate values $\boldsymbol{\Sigma}_j^* = D_j^{*1/2} R_j^* D_j^{*1/2} \sim W(m_0, \boldsymbol{\Sigma}_j)$, $D_j^*$ is a diagonal matrix without restrictions. (3) Accept the new values (replacing $\{D_j^*\}_{11} = 1$) with probability $\alpha = \min\left\{1, \frac{p(R_j^*, D_j^*|-)}{p(R_j^{(t)}, D_j^{(t)}|-)} \frac{q(\boldsymbol{\Sigma}_j|\boldsymbol{\Sigma}_j^*)}{q(\boldsymbol{\Sigma}_j^*|\boldsymbol{\Sigma}_j)}\right\}$, where $p(R_j, D_j|-)$ is the joint posterior distribution and $q(\cdot|\boldsymbol{\Sigma}_j)$ is the proposal distribution given by the product of the Jacobian term for the transformation from $\boldsymbol{\Sigma}_j$ to $(R_j, D_j)$ and the Wishart density $W(m, \boldsymbol{\Sigma}_j)$, such that $J_{\boldsymbol{\Sigma}_j \to R_j, D_j} = \prod_{l=1}^{q_j-1} d_l^{\frac{q_{j-2}}{2}}$.

## 5. Applications

### 5.1. Analysis of the animals dataset

We first test the performance of the proposed model on the animals dataset (Kok & Domingos, 2007; Sutskever et al., 2009). This consists of 50 animal classes and 85 binary attributes (with no missing data). Note that in this experiment we only have a categorical (binary) observation matrix $\boldsymbol{X} \in \{0,1\}^{50 \times 85}$.

The model is fitted using the proposed MCMC scheme. We ran the algorithm considering 20,000 iterations with a burn-in of 5,000 draws, and we collect every third sample that give us a total of 5,000 saved samples. The analysis was performed with $K = 20$, $c = d = 1$, and $\sigma_\lambda^2 = 1$ (many other similar settings yielded similar results). For the sparse probit factor model (discussed in Section 3) we consider six factors; larger models are possible and were considered, how-ever in our experiments we noticed that less than six factors were enough to capture the underlying correlation structure of the binary features. Indeed, only three and four of the columns of $\boldsymbol{B}_r$ and $\boldsymbol{B}_d$, respectively, have non zero elements. In general, the results are very insensitive to the setting of $K$, as long as it is set relatively large.

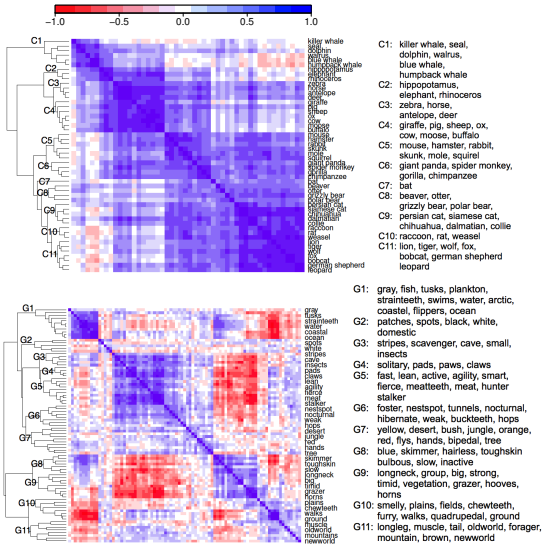

*Figure 1.* Animals dataset: Learned correlation matrices of the distributed representations of the animals (top) and attributes (bottom) as well as their derived dendrograms. Example inferred clusters are denoted by C$_m$ ($m = 1, \ldots, 11$) and G$_n$ ($n = 1, \ldots, 11$) for the animals classes and their attributes, respectively.

We examine the latent correlations (between the rows and columns) learned by the model by inspecting the most likely sample produced by the Gibbs sampler. Figure 1 shows the latent correlation structure between $\{\boldsymbol{r}_i\}_{i=1,\ldots,50}$ learned for the animals as well as between $\{\boldsymbol{d}_j\}_{j=1,\ldots,85}$ learned for the attributes. By the analysis of those correlations, we are able to identify hierarchical clusters and affinities between rows and columns; we use a clustering algorithm (Kaufman & Rousseeuw, 1990) to identify row and column hierarchical cluster structure based on the inferred correlation matrix (this is done for illustration; we do not perform clustering when implementing the model, rather the full covariance between rows and columns is inferred). The hierarchical clustering algorithm yields a dendrogram, plotted jointly with the learned correlation matrices in Figure 1. The closeness of any two clusters is determined by a dissimilarity matrix $\boldsymbol{I} - \boldsymbol{R}$ where $\boldsymbol{R}$ is the correlation matrix (see Eisen et al., 1998; Wilkinson & Friendly, 2009, for more details). The learned groups are described on the right panel of the figure. Some interesting interpretations are derived from the correlation structure for the attributes.

For example, cluster G1 (which includes attributes of marine animals) is highly correlated to cluster G8 and negative correlated to cluster G10 and G11 (which includes attributes like quadrupedal, ground and mountain).

## 5.2. Senate voting data

We next examine a binary vote matrix from the United States Senate during the 110th Congress, from January 3, 2007 to January 3, 2009. The binary matrix, $\boldsymbol{X}$, has dimension $102 \times 657$, where each row corresponds to a senator and each column corresponds to a piece of legislation; $\boldsymbol{X}$ is manifested by mapping all "yes" votes to one and "no" votes (or abstentions) to zero. The percentage of missing values is about 7.1%. We perform analysis of the voting data considering $K = 50$. We use the same priors and MCMC setup considered in the previous application. We inferred that there are approximately 10 binary features for the senators, 13 for the legislation, and $\boldsymbol{M}^{(X)}$ had a rank of $K_x \approx 4$, with one dominant factor, with dominant corresponding $\lambda_t^{(X)}$ (consistent with related research that indicates one dominant factor for such data (Wang et al., 2010)). Figure 2 shows the dendrogram derived from the correlation matrix associated with the senators, to illustrate the clustering of people. The correlation matrix reveals significant differences between two groups of senators, which are constituted by Democrats and Republicans. As an example interpretation of the dendrogram for the senators, note that Republican senators Collins, Snowe and Specter are inferred as being at the "edge" of the Republican party, nearest to the Democrats; these were the *only* Republicans who joined most Democrats in voting for the 2009 Economic Stimulus Package (and Specter later switched to the Democratic party). Also, Barack Obama and Hillary Clinton, who competed closely for the Democratic presidential nomination in 2008, are very weakly correlated with any of the Republicans.

In Figure 2, middle panel, we show the reordered voting data matrix $\boldsymbol{X}$. The matrix was reordered by rows and columns according to similarities learned from the correlations matrices associated with the senators and legislation. The matrix reveals interesting patterns. For example, the first 300 columns are primarily Democrat-sponsored legislation, the following 200 legislation are primarily Republican-sponsored legislation, and the last columns are unanimous votes, for things like nominations for various government posts. We performed **LDA** (Blei et al., 2003) topic modeling on the text documents (separate analysis), to infer structure in the legislation, and help interpret the inferred relationships; three types of legislation so in-ferred are shown in Figure 2.

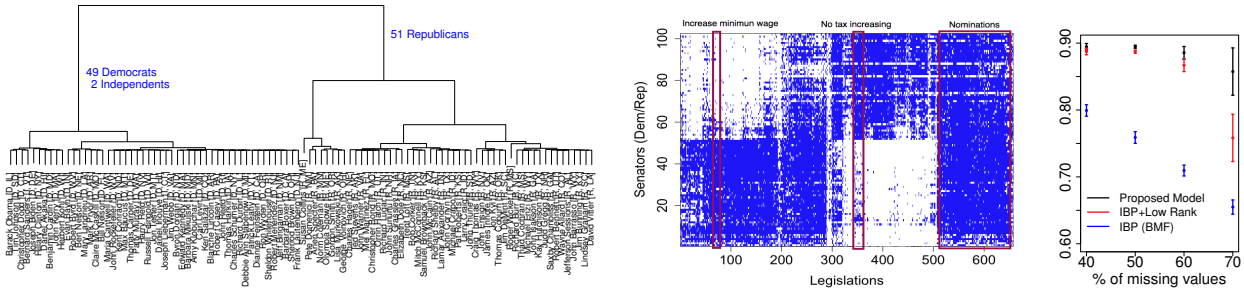

*Figure 2.* Voting data. **Left**: Learned senator groups derived from the correlation structure. **Middle**: Original voting data ("yes" blue, white "no") reordered based on the model; Republicans are at the top-half rows, and Democrats are in the bottom half. Three types of legislation are inferred (rectangles at right), based upon topic modeling (Blei et al., 2003) applied separately to the text legislation. **Right**: Proportion of correct predictions for different fractions of missing data. Error bars indicate the standard deviation around the mean.

We compare our results with two related models. The first follows the proposed construction, except that the latent binary vectors are modeled via an IBP; the second is the logistic *binary matrix factorization* (BMF) model (Meeds et al., 2007); the main difference between the first and second alternatives is that the former imposes the low-rank model of Section 2.4. Figure 2, right panel, shows the average fraction of correct predictions for each model as a function of the fraction of missing data (held-out data, averaged over 15 runs). These results reveal the advantage of the low-rank construction (by comparing the two IBP solutions), and of the imposition of correlation in the latent binary features (omitted in the two IBP-based constructions).

### 5.3. Behavioral dataset

The behavioral dataset comes from a survey conducted by the Duke Visual Cognition Lab during 2010 and 2011 (details omitted here to keep authors anonymous during review). The 508 responders were members of a university community, answering different types of questionnaires; the questions regarded media multitasking (MMI), an attention deficit hyperactivity disorder (ADHD) test, the Autism Spectrum Quotient, eating attitudes (EAT) test, video games (VG) activities, a NEO-AC personality inventory (neuroticism, extraversion, openness, agreeableness, conscientiousness scores) and Barratt Impulsivity Scale (BIS-11); almost all of these questions come from standard surveys in the respective areas (discussed further below). The total dataset consists of $M_1 = 20$ categorical and $M_2 = 106$ real-valued questions. Among the 20 categorical variables considered in the analysis, there are 16 binary observations and 4 variables with more than 2 nominal categories. Concerning the real-valued observations, the 106 studied variables were classified as follows: 40 variables related to VG-playing habits, 23 variables related to passtime activities, 30 associated with the NEO-AC facets, 5 autism subscales, 3 impulsivity subscales, and the last 5 variables related to EAT score, MMI, age, years of education and ADHD score. The percentage of missing values is approximately 13%.

We perform a joint analysis of the categorical and real-valued data matrices considering $K = 50$. The real-valued data matrix $\boldsymbol{Y}$ was column-normalized to zero mean and unit variance before the analysis. In addition, $\sigma_\lambda^2 = 10$, $m_0 = 8$, $\Omega = \boldsymbol{I}_{q_j-1}$ and $c = d = 1$. The MCMC algorithm was run for 50,000 iterations, with the first 25,000 discarded, and then every 5th collected to produce a posterior sample of size 5,000.

Figure 3 shows the approximate posterior distribution for the number of features associated with questions (categorical and real-valued answers) and people. From these results we note that approximately 8 and 6 features in $\boldsymbol{d}_j^{(p)}$ and $\boldsymbol{c}_j$ are used by the model, while there are approximately 20 binary features inferred as associated with the people.

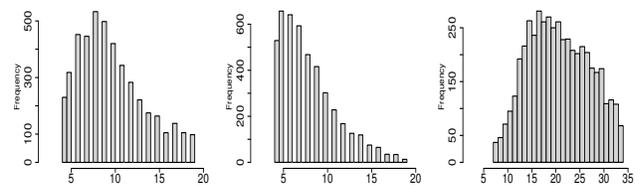

*Figure 3.* Behavioral dataset: Histograms showing the approximate posterior distribution of the number of features associated with questions with categorical answers (left), real-valued answers (middle), and for the people (right).

In a similar fashion to the analysis of the Animals dataset, we analyze the learned correlation matrices associated with the questions and people. Figure 4 shows those matrices with the rows and columns or-dered such that similar rows and columns are near each other. In the vertical margin appears the hierarchical cluster tree derived from the correlation (as discussed above). Based upon these results, we are able to identify blocks of correlated questions and clusters of people. Interesting interpretations can be derived from these results. For example, men are highly correlated with fighting and real time strategy VG and negatively correlated with normal vision and monolingual. Figure 4, right panel, displays the learned correlation structure between people. It shows three big clusters; G1 is primarily composed of women (82.3% women, 17.7% men), G2 represents a heterogeneous-gender group (59% women, 41% men), and G3 is predominantly men (20.6% women, 79.4% men).

We are also interested in the analysis of questions related to behavior scores like the NEO-AC characteristics, autism, ADHD, EAT and MMI. The analysis of these variables is of particular interested in Psychology, where the *Big Five factors* of personality traits (McCrae & John, 1992; Costa & McCrae, 1992) has emerged as a robust model to describe human personality. Specifically, the five factors are directly related with the NEO-AC data (real, non-categorical answers, represented in $\boldsymbol{Y}$) and we seek to connect our inferred latent features with what is known from Psychology, to analyze the quality of inferred structure. From (6) and (8) we have that $y_{ij} = \sum_{l \in \mathcal{S}} \lambda_l^{(Y)} < \boldsymbol{r}_i, \boldsymbol{u}_l^{(Y)} > < \boldsymbol{c}_j, \boldsymbol{v}_l^{(Y)} > + \varepsilon_{ij}$, and therefore from (2) we may express the $l$th component of $\boldsymbol{b}_j$ as $b_{jl} = \sqrt{\lambda_l^{(Y)}} < \boldsymbol{c}_j, \boldsymbol{v}_l^{(Y)} >$, this corresponding to the factor loading for question $j$, factor $l$. Considering the most likely sample in the Gibbs sampling, we infer $\boldsymbol{b}_j \in \mathbb{R}^6$, with $K_y = 6$ factors coming from the rank of $\boldsymbol{M}^{(Y)}$.

Figure 5 shows a diagram where groups of questions are associated to the 6 inferred factors. The plot shows connections between factors and questions in terms of the major values on each factor loading. An interesting finding is that the model *uncovered* the proper number of factors, *i.e.*, five factors that group thirty facets of personality and an additional factor that groups autism scores (to our knowledge, this is the first quantitative analysis of this sort that demonstrates that the question in these questionnaires indeed capture the aspects of personality intended by subject-matter experts in their design). The first five features are clearly related to personality traits, each of them involving different facets of neuroticism (N), extraversion (E), openness (O), agreeableness (A) and conscientiousness (C). Autism scores like communication, social skill and imagination form an additional independent factor. Also, impulsivity scores belong to the factor associated with the *conscientiousness* characteristic but with negative values.

### 5.4. Computations

The code for these experiments was written in Matlab, and the computations were run on a computer with 2.53GHz processor and 4GB memory. To give a sense of computation times, for the Behavioral dataset considered above, approximately 11 seconds were required per MCMC sample.

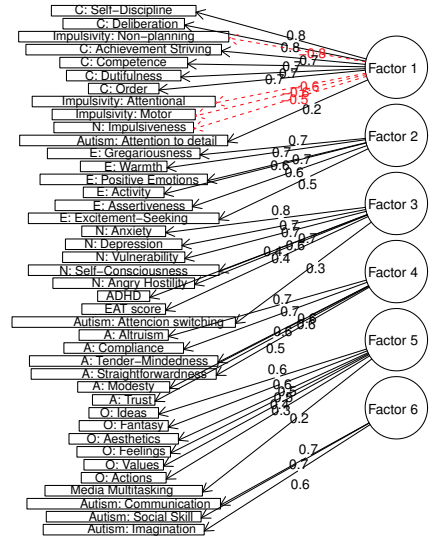

Figure 5. Behavioral dataset: Learned factors that group 30 personality facets of personality (NEO-AC characteristics), autism, ADHD, EAT and MMI, corresponding here to the inferred $K_y = 6$ non-zero components in each $\boldsymbol{b}_j$, for question $j$. At left are types of questions, with significant loading values linked to the factors. The numbers represent the value of the factor loading on the respective question, with negative values showed in red.

### 6. Summary

A new model has been developed for representing real, categorical and mixed real-categorical relational data. A *multivariate* probit model was employed, jointly imposing correlation between the subjects and between the attributes. These covariances were used in the experiments to infer hierarchical structure between the subjects and between the attributes. Encouraging results were demonstrated on three real-world data sets, the last of which is new, characterized by mixed real-categorical survey data for several interesting psychological conditions.

### Acknowledgements

The research reported here was supported by ARO, ONR and DARPA (under the MSEE Program).

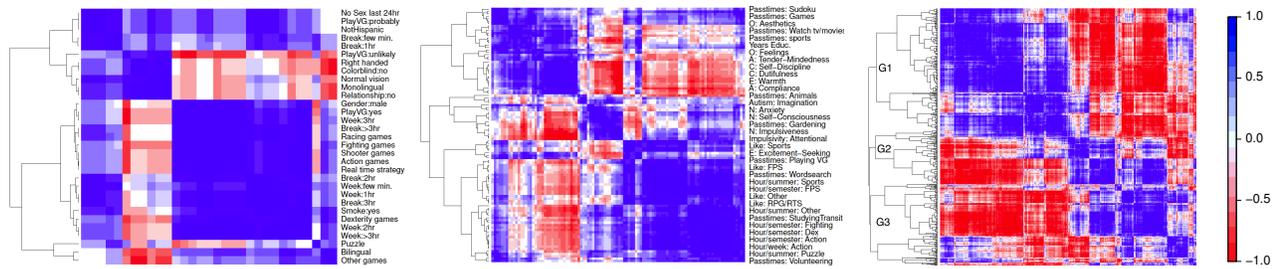

*Figure 4.* Behavioral dataset: Correlation matrices of the answers to the categorical questions (left), real questions (middle) and for people (right). The matrices are plotted jointly with the dendrograms derived from the correlation matrices. In left and middle figures, only a subset of the questions are shown, due to limited space.